\documentclass[lettersize,journal]{IEEEtran}
\usepackage[numbers,sort&compress]{natbib}

\usepackage{amsmath}
\usepackage{array,diagbox,siunitx}

\usepackage{pdflscape}

\usepackage{graphicx}
\usepackage{subcaption}
\usepackage{epstopdf}
\usepackage{subfig}
\usepackage{caption}

\usepackage{cases}
\usepackage{float}
\usepackage{amssymb}
\usepackage{multirow}
\usepackage{mathrsfs}
\usepackage{algorithmic}
\usepackage{rotating}
\usepackage{balance}
\usepackage{threeparttable}
\usepackage{natbib}
\usepackage{color}
\usepackage{colortbl}
\usepackage{makecell}

\definecolor{mygray}{gray}{.9}
\usepackage{makecell}
\usepackage{pdflscape}
\usepackage{siunitx}
\usepackage[pagewise]{lineno}
\usepackage{stfloats}
\usepackage{verbatim}
\usepackage{setspace}
\usepackage{threeparttable}
\usepackage{supertabular,booktabs}
\usepackage{rotating}
\usepackage{supertabular}
\usepackage{pifont}
\usepackage{xcolor,pict2e}
\usepackage{algorithm}
\usepackage{algorithmic}
\usepackage{tikz}
\usepackage{booktabs}
\usepackage{footnote}
\usepackage[pagewise]{lineno}
\usepackage{balance}

\begin{document}
	
	\title{HABD: a houma alliance book ancient handwritten character recognition database}
	
	\author{Xiaoyu Yuan, 
		Xiaohua Huang,
		Zhibo Zhang,
		Yabo Sun
		
		\thanks{X. Yuan, X. Huang, Z. Zhang and Y. Sun are with the School of Computer Engineering, Nanjing Institute of Technology, China}
		
		\thanks{X. Yuan. Xu is with Department of Computer Science, University of Helsinki, Finland}

	}
	
	\markboth{Journal of \LaTeX\ Class Files,~Vol.~14, No.~8, August~2021}%
	{Shell \MakeLowercase{\textit{et al.}}: A Sample Article Using IEEEtran.cls for IEEE Journals}
	
	\maketitle

\begin{abstract}
	
The Houma Alliance Book, one of history's earliest calligraphic examples, was unearthed in the 1970s. These artifacts were meticulously organized, reproduced, and copied by the Shanxi Provincial Institute of Cultural Relics. However, because of their ancient origins and severe ink erosion, identifying characters in the Houma Alliance Book is challenging, necessitating the use of digital technology. In this paper, we propose a new ancient handwritten character recognition database for the Houma alliance book, along with a novel benchmark based on deep learning architectures. More specifically, a collection of 26,732 characters samples from the Houma Alliance Book were gathered, encompassing 327 different types of ancient characters through iterative annotation. Furthermore, benchmark algorithms were proposed by combining four deep neural network classifiers with two data augmentation methods. This research provides valuable resources and technical support for further studies on the Houma Alliance Book and other ancient characters. This contributes to our understanding of ancient culture and history, as well as the preservation and inheritance of humanity's cultural heritage.
\end{abstract}



\section{Introduction}
\label{sec:intro}

As one of China’s precious documents in ancient cultural heritage, the Houma Alliance Book embodies the history and culture of ancient China~\cite{guo2015-1}. Due to the extensive use of ancient characters in the Houma Alliance Book, traditional human-level  explanations are no longer sufficient to meet the demands for cultural dissemination.  

Research on the characters found in the Houma Alliance Book dates back to the 1960s. This ancient document was inscribed during the late Spring and Autumn period, specifically during the 15th to 23rd year of Duke Ding of Jin's reign (497-489 BC), serving as a convenant between Zhao Yang, the chief minister of the Jin state, and other ministers. In 1965, during the excavation of the Houma Jin State site coducted by the Shanxi Provincial Cultural Relics Work Committee from November of that year to May of the following year, the Houma Alliance Book was discovered. This remarkable find not only advanced the study of Chinese characters but also filled a crucial gap in textual materials. 

The Houma Alliance Book deepened our understanding of covenant methods during the Spring and Autumn to the Warring States period. Scholars such as~\cite{guo2015-1} have meticulously organized and studied these ancient texts. In the early 1970s, the Shanxi Provincial Cultural Relics Work Committee spearheaded comprehensive research efforts on these materials, which included copying, organizing, and analysis. Their findings culminated in the publication of the ``Houma Alliance Book: Archaeological Report" by~\cite{Zhang1976}. This work servers as a valuable source of research materials for the creation of the Houma Alliance Book ancient handwritten character database.

Currently housed as one of the prized cultural relics at the Shanxi Museum, the Houma Alliance Book stands as a crucial source of information for the  study of Eastern Zhou characters. Within the realm of Chinese calligraphy history research, these artifacts unquestionably represent invaluable physical materials. Nevertheless, owing to their ancient origins, the extant related literature is incomplete and scarce. Furthermore, both domestically and internationally, there is a notable absence of research on the recognition of the Houma Alliance Book, coupled with the varying character forms within the slips. These factors undeniably amplify the challenges associated with manual recognition.

As a living specimen of ancient Chinese calligraphy, the Houma Alliance Book meticulously records numerous historical events from the Spring and Autumn period, rendering it of exceptional important for understanding and interpreting their content. Nonetheless, the development of databases for the ancient characters within the Houma Alliance Book remains an unresolved issue. 


With the continuous advancement of deep learning technology and the emergence of numerous excellent models, deep neural networks have demonstrated tremendous advantages across various domains, including speech recognition, image recognition, and natural language processing. Leveraging deep learning presents a promising avenue for ancient handwritten character recognition in Houma Alliance Book. However, one critical challenge is the scarcity of data, which may limit the application of deep learning techniques. Our research findings indicate that the data distribution within the Houma Alliance Book ancient handwritten character database follows a long-tail pattern. This characteristic introduces unique challenges for recognition. While deep neural network models excel in feature extraction and classification, they must address the long-tail nature of the data distribution to achieve efficient and accurate character recognition. The long-tail distribution suggests that the majority of ancient characters have low occurrence frequencies, with only a few appearing frequently. Consequently, an imbalanced data distribution emerges. In traditional deep neural network models, data imbalance can lead to overfitting on low-frequency data and underfitting on high-frequency data, which ultimately impacts recognition accuracy.

In this paper, we introduce a new database for recognizing ancient handwritten characters from Houma Alliance Book, along with a benchmark designed to address issues related to data scarcity and long-tail distribution of characters. Specifically, firstly, we employ an iterative approach to collect and annotate the houma alliance book character from the Houma Alliance Book, resulting in a new database comprising 26,732 images and 327 classes. Secondly, we enhance recognition accuracy by applying long-tail techniques, such as Mixup, to balance the data distribution among different categories. Lastly, to leverage the advantages of various deep neural networks, we implement decision-level classifier fusion across four architectural models, thereby establishing a substantial baseline for character recognition within the Houma Alliance Book. 

The main contributions of this work are summarized as follows:
\begin{itemize}
	 \item We propose a new Houma alliance book handwritten character database by leveraging data-acquisition techniques to tackle the ancient character recognition problem in the Houma Alliance Book.  
	
	\item Based on the new database, we introduce an benchmark as our baseline. The baseline algorithm incorporates long-tail learning method and classifier fusion across four deep learning architectures.
	
	\item This research provides valuable resources and technical support for further studies on the Houma Alliance Book and other ancient characters. contributing to our understanding of ancient culture, history and the preservation of humanity's cultural heritage.
\end{itemize}


The rest of this paper is organized as follows: Section~\ref{sec:related} reviews the related work and research in relevant fields. Section~\ref{sec:database} provides details about the Houma Alliance Book ancient handwritten character database, including its structure, scale, and characteristics. Section~\ref{sec:method} introduces the benchmark methods for this database. Section~\ref{sec:exp} presents the experimental analysis of the new database. Finally, a conclusion is given in Section~\ref{sec:conclusion}. Preliminary results of part of this work have been published in~\cite{Yuan2023}.

\section{Related work}
\label{sec:related}


\subsection{Ancient handwritten character database}

With the advancement of image processing technology, researchers have established numerous databases cofused on historical handwritten character. ~\cite{liu2013online} introduced the Chinese Handwriting Database (CASIA-HWDB), which includes both online and offline versions. This extensive database comprises Chinese characters handwritten by 1,020 individuals, including 7,185 isolated Chinese characters, 171 symbols, and 5,090 pages of text, totaling approximately 1.35 million characters. The study primarily employed traditional machine learning methods, utilizing features such as grayscale and stroke trajectory during feature extraction. Classifiers such as quadratic discriminant functions, discriminative feature extraction, and discriminative learning quadratic discriminant functions were chosen, achieving a maximum accuracy of 93.22\%. Additionally, ~\cite{zoizou2022moj} introduced a contour-based feature extraction method and developed the Arabic Historical Words Database (MOJ-DB). The study utilized 64 pages of Arabic historical documents, which contained a total of 5,600 subwords. The database was created by segmenting these subwords using contour feature extraction. Furthermore, the research employed a convolutional neural network model, HOG+EHD for text feature extraction, and support vector machines as classifiers, achieving recognition results of 89.6\% and 83.4\% respectively. 

\subsection{Long-Tail Methods in Deep Neural Networks}

Research has identified the long-tail problem as one of the most challenging issues in visual recognition (\cite{zhang2023deep}). The long-tail phenomenon occurs when data is distributed in such as way that a few categories have a large number of samples, while the remaining categories have relatively few samples. Categories with a significant number of samples are often referred to as `head' categories, whereas those with fewer samples are termed `tail' categories. Common long-tail datasets in research include iNaturalist 2018 (\cite{van2018inaturalist}), ImageNet-LT (\cite{liu2019large}), CIFAR100-LT (\cite{cao2019learning}), and Places-LT (\cite{liu2019large}). Among these datasets, iNaturalist 2018 is considered a genuine long-tail dataset, while the other three are long-tail datasets created by sampling from larger datasets like ImageNet, CIFAR100, and Places, following the Pareto distribution. The sample data of the ancient characters from the Houma Alliance Book in our study exhibit a similar long-tail distribution characteristic to that of iNaturalist 2018, As shown in Fig.~\ref{fig:class} further discussed in Section~\ref{sec:database}, there are fewer categories in the `head', each containing a substantial number of ancient character samples, while the `tail' comprises numerous categories, each with a smaller number of ancient character samples.

Within the context of a long-tail distribution, sample features are heavily influenced by the head categories, making it challenging for deep neural network algorithms to optimize and avoid overfitting on the tail categories. Consequently, the adoption of various long-tail approaches becomes imperative to enhance the algorithmic efficiency in recognizing the tail categories. In a recent study by~\cite{zhang2023deep}, the main issues surrounding long-tail recognition were discussed, and they summarized  categories and significant applications of existing long-tail methods, offering valuable insights for future research. This study categorized long-tail methods into three types: class balancing, information enhancement, and module improvement, providing an overview of long-tail algorithm research, according to~\cite{liu2019large}. 

Class balancing techniques aim to address differences in the number of samples across categories from the input layer to the loss layer. Such methods include class resampling, class cost-sensitive learning, and log-odds adjustment. 

Information enhancement can be categories into two main techniques: transfer learning and data augmentation. These methods emphasize the introduction of external information to enhance the recognition performance of the tail categories while preserving the recognition accuracy of head categories. For example, the Mixup method proposed by~\cite{zhang2017mixup} randomly selects two images and labels from each batch and generates new images and labels for training through random linear interpolation, effectively achieving data augmentation. This approach has demonstrated improved accuracy, particularly on datasets with small sample size, thereby validating its feasibility.

Module improvement seeks to enhance the recognition performance of long-tail data from a classifier perspective, encompassing aspects such as feature extractors, classifiers, decoupled training, and ensemble learning. ~\cite{cui2022reslt} argued that traditional methods like resampling or reweighting might lead to overfitting on tail categories and may not fundamentally solve the data imbalance issue inherent in long-tail distribution. They proposed a long-tail method based on the residual learning mechanism, which combines three training branches in the parameter space and established a residual fusion module to optimize long-tail algorithms.~\cite{zhang2021bag} conducted extensive experiments and analysis on long-tail algorithms. They utilized ImageNet-LT as a benchmark dataset and performed experiments with various long-tail techniques, including resampling, reweighting, Mixup, and Class Active Mapping, achieving noteworthy results. These studies provide a robust theoretical and experimental foundation for addressing the long-tail problem in recognizing the ancient handwritten characters of the Houma Alliance Book.


\section{Houma Alliance Book Handwritten Character Recognition Database}
\label{sec:database}

This section introduces the structure and characteristics of the Houma Alliance Book ancient handwritten character database.

\subsection{Iterative Database Collection Method}

To ensure annotations consistency, we construct the Houma Alliance Book ancient character database (HABD) using an iterative data collection and annotation method. Due to the Houma Alliance Book's extensive history, the data samples contain high levels of noise, making it challenging to directly acquire sufficient samples from the original data. The following factors contribute to this issue:
\begin{itemize}
	\item Limited number of annotated ancient character samples.
	\item The ancient character styles during the pre-Qin period were diverse, lacking a unified style. This leads to multiple possible variations for a single ancient character, increasing the reliance on training data.
	\item  Previous research on the Houma Alliance Book provided a large number of unlabeled ancient character samples, requiring manual annotation or the use of semi-supervised methods to be used as training sets.
\end{itemize}

\begin{figure}[t!]
	\centering
	\includegraphics[width=\linewidth]{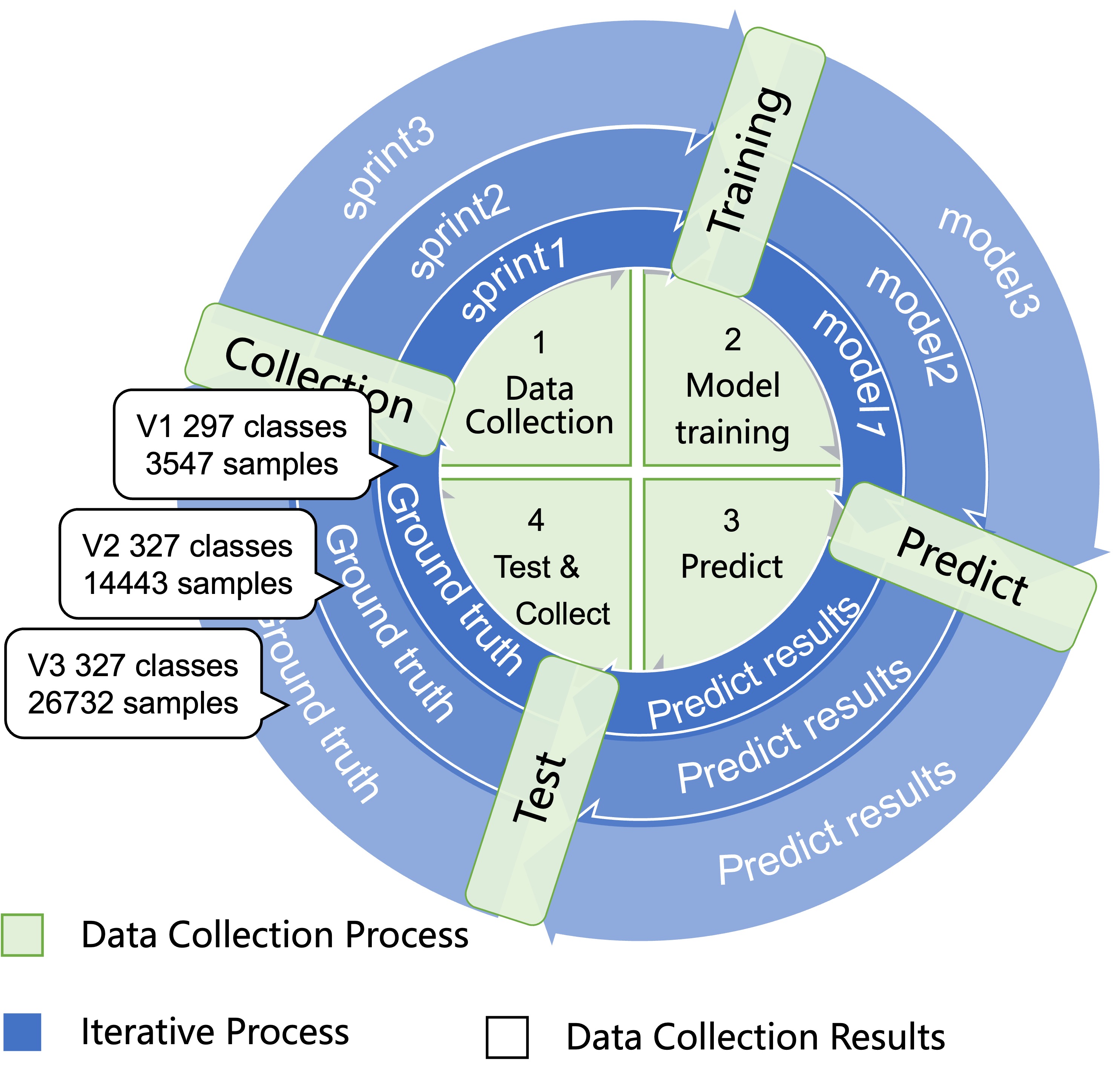}
	\caption{Data collection process for HADB.}
	\label{fig:fig1}
\end{figure}

Given the characteristics mentioned above, we have adopted an iterative data collection and annotation method for recognizing and annotating the existing ancient characters within the Houma Alliance Book. At the heart of this method lies a cyclic process consisting of ``data collection, model training, model prediction, test verification, and dataset update". Through multiple iterations, this process acquires a sufficient number of characters. As depicted in Fig.~\ref{fig:fig1}, the data collection process underwent three iterations.

\subsection{Acquisition details}

We obtained identified ancient handwritten characters and their corresponding traditional Chinese characters from the archaeological report of the Houma Alliance Book~\cite{Cheng2002}. Initially, we collected and organized a small dataset comprising 327 identified character classes, each with multiple variants. A total of 14,443 original images are extracted from the archaeological report of the Houma Alliance Book, with the existing copies of the archaeological report cropped to a size of 64×64 pixels. 

As the data collection was based on the 327 identified ancient character classes mentioned in the archaeological, and as the identification results mainly corresponded to traditional Chinese characters, some of which are now less common or nonexistent, we utilized a specialized character compilation program to create data labels by constructing characters. More specifically, we utilized an iPad digital handwriting tool and engaged four individuals to mimic the Houma Alliance Book's ancient handwritten characters. 

\begin{figure}[t!]
	\centering
	\includegraphics[width=\linewidth]{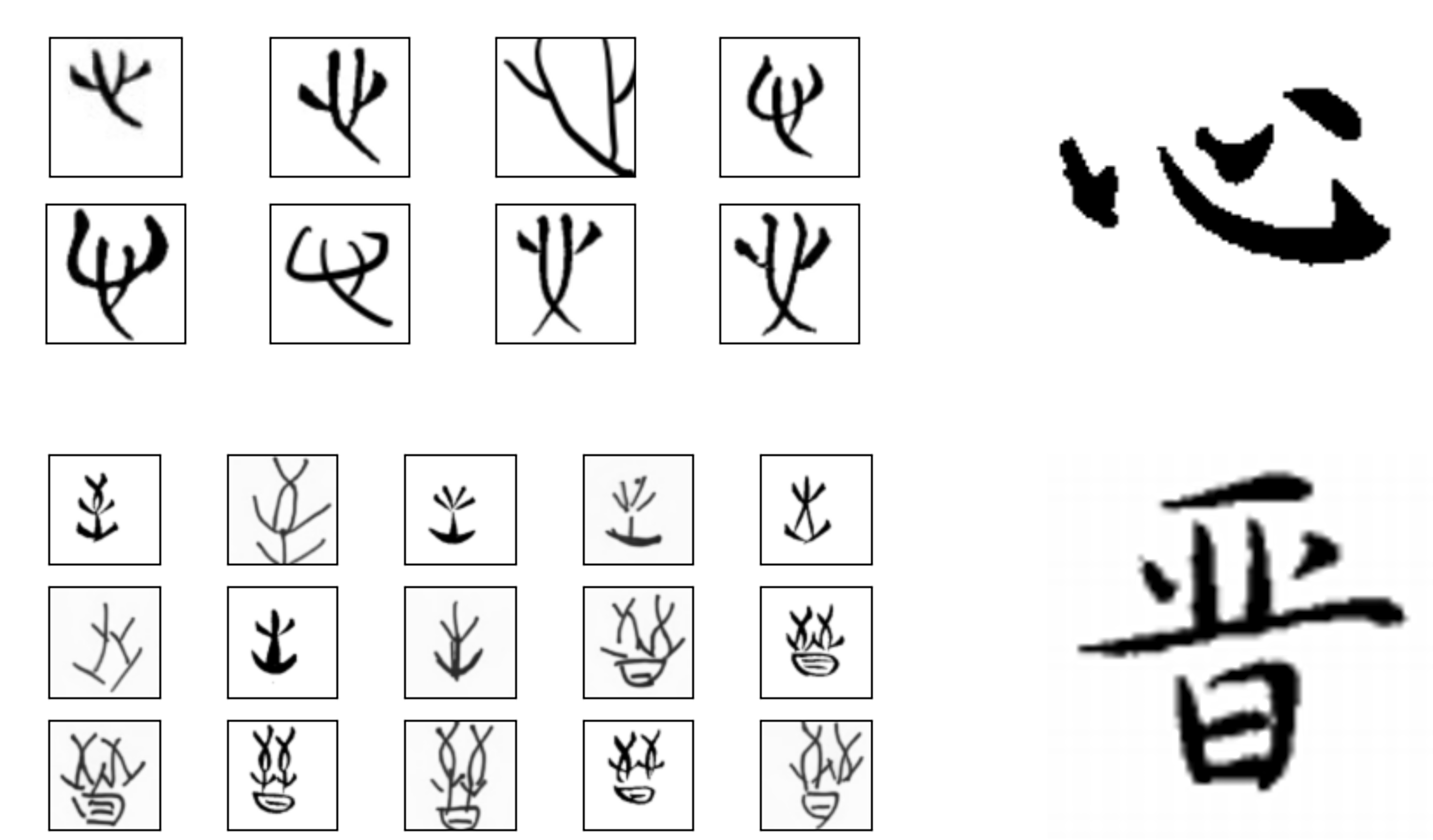}
	\caption{Two examples in the new database, where the left columns are Houma Alliance Book ancient handwritten characters, and the last column is their corresponding annotated traditional Chinese character.  }
	\label{fig:sample}
\end{figure}

\begin{figure*}[t!]
	\centering
	\includegraphics[width=\linewidth]{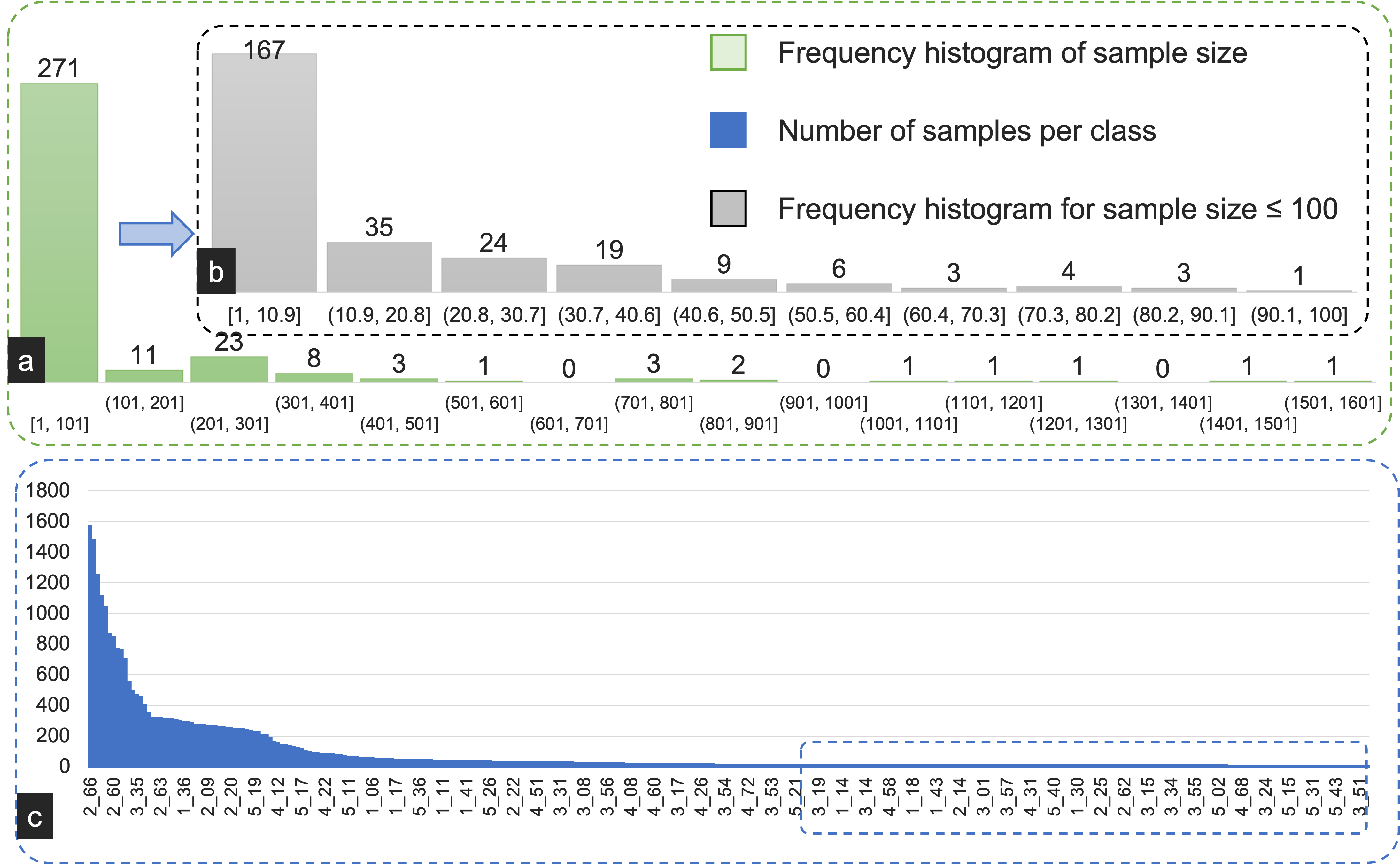}
	\caption{Class distribution in the database. (a) Frequency distribution histogram of sample numbers (b) Frequency distribution histogram of categories with sample numbers not exceeding 100 (c) Sample numbers for each category, exhibiting an obvious long-tail phenomenon. }
	\label{fig:class}
\end{figure*}

Through this annotation process, we create a database comprising 26,732 images, each annotated with the corresponding traditional Chinese characters. Examples from the dataset and its data distribution are shown in Fig.~\ref{fig:sample} and Fig.~\ref{fig:class}. As illustrated in Figure~\ref{fig:class}(a), sample numbers for 271 categories fall within the range of [1,101], with 167 categories having fewer than 10 samples, as depicted in the frequency distribution histogram in Figure~\ref{fig:class}(b). The overall sample distribution across categories is presented in Figure~\ref{fig:class}(c).

During the database development process, two significant issues emerged. Firstly, some ancient characters displayed mirror symmetry, which hindered system performance when using horizontal mirroring for data augmentation. To address this, we opted for vertical mirroring and random angle flipping during data preprocessing, methods more suitable for ancient character recognition. Secondly, certain ancient characters exhibited similar features with subtle differences such as variations in radical components or stroke reduction. These nuances posed challenges to the effectiveness  effectiveness of deep learning models.

\section{Benchmark for HABD}
\label{sec:method}

\subsection{Overview}
To address the issue of long-tail data, we adopt the Mixup data augmentation method, which effectively improves the performance and accuracy of ancient character recognition. Furthermore, we introduce four deep network architectures, including AlexNet, ResNet, Vision Transformer (ViT), and Cross Transformer models (CrossViT). Through deep learning techniques, these classifiers accurately classify and recognize ancient characters. Finally, we propose a decision-level fusion model that combines the results of multiple classifiers to enhance the accuracy and robustness of ancient character recognition. By applying these key techniques and methods, we achieve accurate and efficient recognition of ancient characters. The algorithm performs data augmentation at the input and loss layers, as shown in Figure 4. The algorithm consists of three parts: input layer data augmentation, classifier training at the training layer, and model fusion at the loss layer.

\subsection{Data augmentation}
In the input layer, the Mixup long-tail optimization method is used to augment the features of the long-tail distribution. In each batch, two samples are randomly selected, and a new sample is generated using random linear interpolation. Let $x_1$ and $x_2$ be the randomly selected sample images, and $y_1$ and $y_2$ be their corresponding labels. Let $\lambda$ be the mixing coefficient, where $\lambda \in [0, 1]$. The new sample $x$ and new label $y$ are defined as:
\begin{equation}
	x=\lambda x_1+(1-\lambda)x_2,
\end{equation}
\begin{equation}
	y=\lambda y_1+(1-\lambda)y_2.
\end{equation}

The generated samples are used for training the deep network, while $x_1$ and $x_2$ are not involved in the training. Since the input layer mixes the images and labels through Mixup~\cite{cao2019learning}, four different deep network classifiers are trained in parallel during training. The loss function $L_n$ for the $n$-th classifier can be expressed as:
\begin{equation}
	L_n = \lambda f(y', y_1) + (1-\lambda)f(y', y_2),
\end{equation}
where $\lambda$ is the mixing coefficient, and $f(y', y_k)$ is a commonly used loss function that calculates the loss between the original label $y_k$ and the predicted label $y'$.

\subsection{Classifier selection and analysis}

This study leverages advanced technologies and assesses the strengths of Convolutional Neural Networks (CNNs) and Transformers in the field of ancient character recognition. Through experimental validation and analysis, four benchmark models including AlexNet (\cite{krizhevsky2012imagenet}), ResNet (\cite{he2016deep}), Vision Transformer (\cite{dosovitskiy2020image}), and CrossViT (\cite{chen2021crossvit}) were chosen for recognizing the Houma Alliance ancient handwritten characters.

AlexNet was the first deep-learning model to achieve top results in the ImageNet challenge. This comprises multiple convolutional layers, pooling layers, and fully connected layers. Convolutional and pooling layers are responsible for feature extraction, while fully connected layers (including classification) are utilized for feature classification. Specifically, AlexNet consists of 5 convolutional layers and 3 fully connected layers. Each convolutional layer is equipped with a ReLU non-linear activation function to enhance the model's non-linear expressive capability. Additionally, AlexNet adopts Dropout techniques to prevent overfitting.

ResNet was one of the first models to address the vanishing gradient problem in deep CNNs, and was among the top-performing models in the ImageNet challenge that year. In conventional deep CNNs, increasing the network depth can lead to gradient vanishing issues and overfitting. To tackle this, ResNet introduced "residual blocks" with skip connections that add the input directly to the output, mitigating gradient vanishing. Moreover, ResNet incorporates batch normalization techniques to further enhance model expressiveness and stability.

Vision Transformer (ViT) is an image classification model based on the Transformer architecture. Unlike traditional CNNs, ViT replaces convolutional operations with self-attention mechanisms, enabling end-to-end training and inference. ViT divides input images into smaller patches and applies linear transformations to obtain vectors representing each patch. These vectors are concatenated and augmented with positional encodings before being input to the Transformer. The self-attention mechanism in the Transformer effectively processes this sequence and outputs a vector representing the entire image for final classification. Multi-head attention, residual connections, and layer normalization components within the self-attention mechanism enhance the model's expressive power while mitigating gradient vanishing and exploding issues. During training, ViT employs the SGD algorithm to optimize the loss function. Dropout and data augmentation techniques are also used to prevent overfitting and expand the training dataset.

Cross Transformer (CrossViT) is a neural network model designed for cross-modal learning. It handles both image and text information simultaneously, aligning and fusing them to enhance cross-modal task performance. Built upon the Transformer architecture, Cross Transformer employs self-attention mechanisms for cross-modal data processing. It includes visual and language modules responsible for processing image and text information, respectively. These inputs are encoded into vectors and then processed within the CrossViT. The self-attention mechanism in the CrossViT comprises multi-head attention, residual connections, and layer normalization. These components enable simultaneous focus on visual and language information, resulting in a vector that represents the alignment and fusion of both modalities for cross-modal predictions. In training, CrossViT adopts joint training, optimizing both visual and language module loss functions simultaneously. Additionally, dropout and data augmentation techniques are used to prevent overfitting and expand the training set.

\subsection{Classifier fusion}

Classifier fusion has been presented on several types of classifiers to enhance the performance of individual classifiers. The previously proposed works~\cite{huang1995method,abdi2018deep} demonstrated more precise, accurate, and certain aspects of any decision that constructs the fusion. Normally, with this context, a set of decisions is first conducted and then synthesized via a particular classifier fusion strategy. Mathematically, given a test image $x$, with $N$ base classifiers, the probability for the $l_n$ class predicted by the nth classifier is denoted as $\theta_(n,l_n)(x)$, where $l_n \in {1,…,C}$, and the conditional independence can be defined by
\begin{align}
		\theta_(n,l_n)(x|\omega)&=\theta_(l_1,\ldots,l_N|x_t,\omega)
		                                         &=\prod_{n=1}^{N}\theta_(n,l_n)(l_n|x,\omega),
\end{align}
where $\omega \in {1,\ldots,C}$ is the class label predicted by the classifier combination. The posterior probability required for the $n$-th classifier can be explained by
\begin{equation}
		\theta_(n,l_n)(x|\omega)=\frac{\theta(x)\theta(x|\omega)}{\theta(\omega)}=\frac{\theta(x)\prod_{n=1}^{N}\theta_(n,l_n)(l_n|x,\omega)}{\theta(\omega)}.
\end{equation}

It is noted that $\theta(x)$ does not reply on $l_n$. Therefore, the final probability of $x$ is formulated as,
\begin{equation}
	\sigma_{\theta}(x)=\max_{l_n}(\omega_k)\prod_{n=1}^{N}\theta_{n,l_n}(\omega|x),
\end{equation}
where the $max$ function is employed to choose the final class w.r.t. the maximum probabilities of all classes. 

In our system, the base classifiers are AlexNet, ResNet, ViT and CrossViT. The voting method obtains the result of the fusion model by training multiple classification models at the same time and merging and averaging. The voting method includes hard voting and soft voting. Hard voting obtains the fusion result by voting on the prediction results of various models while soft voting uses the possibility predicted by various models as weights to perform weighted voting based on hard voting. Generally, soft voting obtains better results than hard voting. Therefore, the soft voting method is used to perform decision-level classifier fusion (DCF) across four base classifiers. Finally, we obtain four ensembled models.

For classifier fusion, four base models are performed by 300 epochs and using a batch size of 64; The learning rate is 0.001. To improve the performance of the three base architectures, data augmentation is introduced. For each original image, we generate a horizontally flipped image. We apply the following five image processing techniques: scaling, random translation, image rotation, and color brightness. In detail, we randomly scale the data with different sizes. Then, we randomly horizontally flip the samples. Additionally, we set up rotation to enhance the data. In addition, we generate different samples with different brightness, contrast and color brightness.

\section{Experiment}
\label{sec:exp}
To validate the performance of the proposed method in the new database, we implement PyTorch for five deep architectures and decision-level classifier fusion. According to class distribution, we randomly select 200 samples for the class that have more than 200 samples. 11,534 are used training, 2761 for testing.  Our system named ``Meet with HMMS'' is online~\footnote{http://43.138.135.148}.

\subsection{Performance comparision}

Four commonly used deep learning models (LeNet (\cite{lecun1989handwritten}), AlexNet, ResNet, Vision Transformer, and CrossViT) were implemented as baseline classifiers for image classification tasks. These models were used to predict the baseline dataset to determine their performance on this task. The recognition results are reported in Table~\ref{tab:perf}.

\begin{table}
	
	\caption{Performance comparision (\%) on three neural networks and two transformer-based networks. Best result is in bold.}
	\begin{tabular}{c|ccccc}
		\hline
		Method & LeNet & AlexNet & ResNet & ViT & CrossViT \\
		\hline
		Accuracy & 57.98 & 87.45 & 88.23 & 90.52 & \textbf{90.72} \\
		\hline
		
	\end{tabular}
\label{tab:perf}
\end{table}

In the recognition of single characters in the Houma Alliance ancient handwritten characters, ResNet, ViT, and CrossViT outperformed the other models. This indicates that compared to convolutional neural networks, residual learning and Transformer models based on attention mechanisms are more suitable for recognizing ancient characters.

\subsection{Long-tail method optimized deep neural network}
\begin{figure}[t!]
	\centering
	\includegraphics[width=\linewidth]{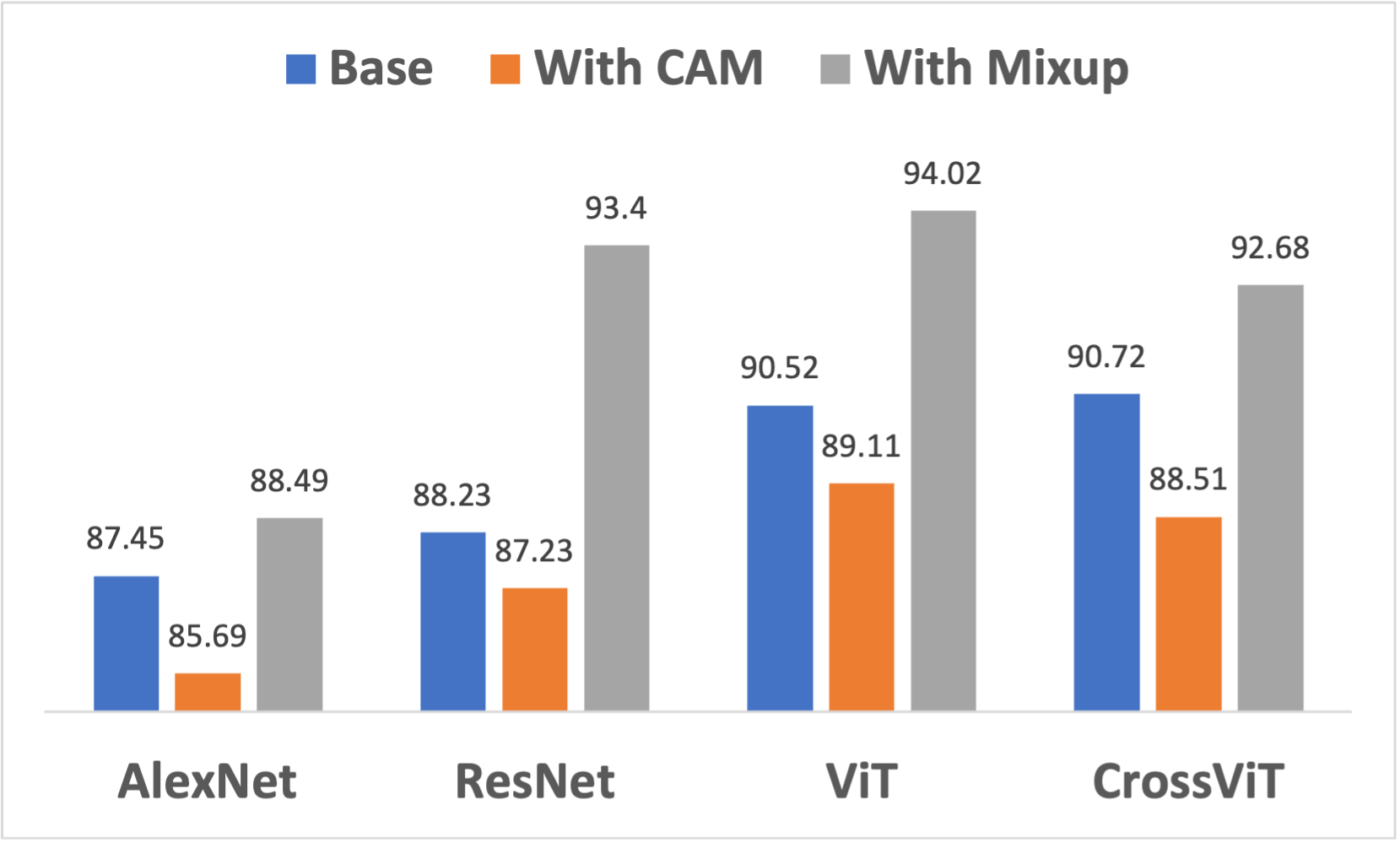}
	\caption{Recognition accuracy (\%) of long-tail methods on baseline classifiers}
	\label{fig:longtail}
\end{figure}

In this study, the long-tail algorithm was incorporated to improve the training and inference processes of the models. Specifically, the Mixup algorithm was used to enhance the model's generalization ability, the CAM algorithm was used to improve the model's interpretability, and the Transform method was used to augment the dataset with vertical mirroring, rotation angles, etc. These baseline models were then retrained, and their predictions on the test set were used to evaluate the contributions of these algorithms and techniques. The recognition results are shown in Fig.~\ref{fig:longtail}.

Since the Houma Alliance ancient handwritten character database constructed in this paper was derived from rubbings by archaeologists, which had minimal noise and standardized images, the CAM algorithm, which is suitable for complex scene feature extraction, did not perform well on this database. On the other hand, the Mixup method effectively balanced the data imbalance issue in the long-tail database and improved the recognition accuracy by 1-4\% across the four baseline classifiers.

\subsection{Long-tail method optimized fusion model}
By combining the long-tail method optimization and decision-level fusion, the Mixup+ResNet, Mixup+CiT, and Mixup+ViT models with higher recognition performance were selected, and a soft voting strategy was used for fusion at the decision level. The fusion results are shown in Table~\ref{tab:fusion}. The DLF-RVC fusion model, optimized based on the long-tail algorithm, achieved a recognition accuracy of up to 95.21\%.

\begin{table*}[t!]
	\caption{Performance comparision (\%) of fusion models optimized by the long-tail method. Best result is in bold.}
	\centering
	\begin{tabular}{c|ccc}
		\hline
		Method & Classifiers & No-optimized & Optimized \\
		\hline
		DLF-RV & ResNet + ViT & 92.74 &93.84\\
		DLF-RC & ResNet + CrossViT & 92.74 &94.32\\
		DLF-VC & ViT + CrossViT & 92.22 &93.93\\
		DLF-RVC & ResNet + ViT + CrossViT & 93.34 &\textbf{95.21}\\
		\hline
		
	\end{tabular}
	\label{tab:fusion}
\end{table*}

\begin{figure*}[t!]
	\centering
	\includegraphics[width=\linewidth]{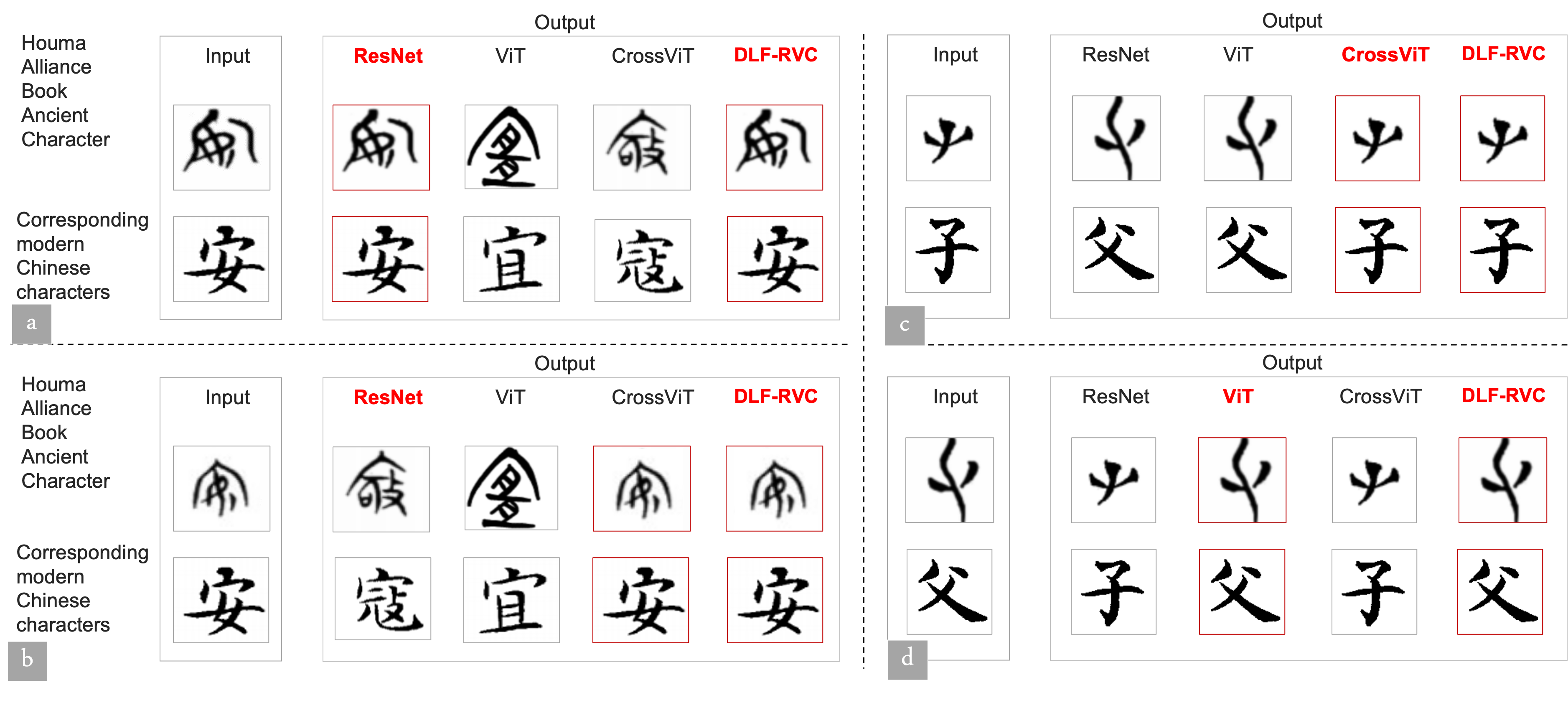}
	\caption{Recognition results on three algorithms and DLF-RVC, where the algorithm that correct predict the character is red in bold. (a) and (b) are test results for different typefaces of the label ancient characters. (c) and (d) are test results for two mirror-image versions of ancient characters.}
	\label{fig:resultshow}
\end{figure*}

Fig.~\ref{fig:resultshow} shows the recognized results by three algorithms and DLF-RVC. It is seen that our DLF-RVC can effectively recognize the difficult type of ancient characters, especially some ancient characters which exhibits mirror symmetry. It considerably demonstrates the generalization ability of DLF-RVC algorithm proposed in this study.

\section{Conclusion}
\label{sec:conclusion}

In this paper, we propose a new Homa Alliance Ancient character database and offer deep neural network algorithms for benchmarking character recognition performance. To address this, various models and feature extraction methods were attempted, and a long-tail approach was adopted to optimize the baseline classifier, thereby enhancing recognition accuracy. In conclusion, our future focus lies in expanding the scope of ancient Chinese character recognition and applying deep learning methods to more intricate historical documents. We believe that this research can contribute to the development of ancient Chinese language studies and offer valuable insights into the realm of Chinese language, literature, and history research.

\ifCLASSOPTIONcaptionsoff
\newpage
\fi

\footnotesize
\bibliographystyle{IEEEtran}
\bibliography{HABD}

\vfill

\end{document}